# Towards Practices for Human-Centered Machine Learning

Stevie Chancellor

University of Minnesota – Twin Cities, steviec@umn.edu

"Human-centered machine learning" (HCML) is a term that describes machine learning that applies to human-focused problems. Although this idea is noteworthy and generates scholarly excitement, scholars and practitioners have struggled to clearly define and implement HCML in computer science. This article proposes practices for human-centered machine learning, an area where studying and designing for social, cultural, and ethical implications are just as important as technical advances in ML. These practices bridge between interdisciplinary perspectives of HCI, AI, and sociotechnical fields, as well as ongoing discourse on this new area. The five practices include ensuring HCML is the appropriate solution space for a problem; conceptualizing problem statements as position statements; moving beyond interaction models to define the human; legitimizing domain contributions; and anticipating sociotechnical failure. I conclude by suggesting how these practices might be implemented in research and practice.

## 1 INTRODUCTION

Machine learning (ML) has been described as a modern Oracle of Delphi—a way to quickly solve problems in different domains, whether auto-completing emails or predicting the presence of malignant tumors. Computer scientists are theorizing and designing ML technology into our social and personal lives. ML has justifiably stirred tremendous excitement in research, industry, and the popular zeitgeist of artificial intelligence (AI).

However, the enthusiastic adoption of ML has also had negative consequences. ML is being used for unsavory and controversial purposes, such as generating "deep fake" videos and reproducing racial discrimination [8,41]. On the research side, new research points to worrisome trends of chasing metrics over more principled approaches and questionable gains in deep learning's performance compared to linear models [10,17,30]. As a researcher working in applied machine learning for mental health, I have seen the consequences of this firsthand, where principled approaches to methods are neglected [13] and people become the objects of prediction in a pipeline [11]. Computer science (CS) is at a stage where interest in ML has both brought energy to new questions as well as raised reasonable concerns.

In response, a new area called "human-centered machine learning" (HCML) promises to balance technological possibilities with human needs and values. Put simply, HCML couples technical innovations in ML with social values like fairness, equality, and justice. The focus for HCML is broad; it includes fair and transparent algorithm design [39], human-in-the-loop decision-making [48], design for human-AI collaborations [4], and exploring the social impacts of ML.

However, there are no unifying guidelines on what "human-centered" means, nor how HCML research and practice should be conducted. People have worked to articulate a nascent set of values for HCML [11,44,47], but the concept is not clear and definitions come from many sides. This moment provides an opportunity to refine HCML into a unifying and interdisciplinary force across CS rather than risk fracture with each sub-field of CS taking ownership of an independent vision of HCML [29].

This article draws on the interdisciplinary history of human-centered thinking, HCI, AI, and Science and Technology Studies (STS) to propose best practices for human-centered machine learning. I also draw from lessons learned from my experiences as a scholar who transitioned from Media Studies to Human-Computer Interaction and applied ML. By adapting these lessons learned and discussing computing's responsibility for society through a sociotechnical lens [43], this article provides a beginning set of HCML practices to help scholars conduct research that is rigorous without looking sight of values and pro-social goals. These practices are intended to be revisited through the whole HCML pipeline of problem brainstorming, development, and deployment. Finally, I conclude with pragmatic examples that extend these practices into disciplinary changes to ensure HCML's future success.

## 2  WHAT IS HUMAN-CENTERED MACHINE LEARNING?

HCML has captured the interest of ML and other computing disciplines. This interest has been reflected in a series of workshops and activities on human-centered perspectives at major HCI venues, such as ACM SIGCHI in 2016 [22], and two in 2019 [26,38]. There has been attention in ML too, as shown through the rapid ascent of the FAT ML workshop to its own independent ACM FAccT conference. The term "human-centric" has also appeared in workshops in NeurIPs[1] and in a special issue in ACM TiiS [20]. These workshops and special issues discuss common themes, such as prioritizing human needs and values like explainability, studying the human labor in ML engineering, and audits of deployment. Collectively, these represent opportunities for humans to interface with ML systems.

I have been working in HCML since 2014, researching how machine learning can detect mental illness expressed through social media data [11,12,14]. This research agenda is complex, with technical challenges around data size and modeling capabilities; challenges of mapping psychological constructs to ambiguous social media data; and social issues of governance, privacy, and equity. At the start of my journey, there were no terms that explained my interest in both ML and social implications. HCML and its umbrella of concern provides a crucial home for my research that jointly considers technical innovation alongside social concerns.

But having a term to define the field in my head has not made it easier to describe to others. People often ask me what "counts" as HCML and interrogate what it means. Does working with human-generated data count?

---

[1] https://sites.google.com/view/hcml-2019

If I do interviews or ask users what they want, does that count? What if I don't build an AI system, but ask people about its impacts? How about if I point out the racial disparity of AI systems but never build anything and don't have a CS degree? Attaining a definition is also complicated because, although HCI and ML are players in this area, other disciplines such as critical data studies and STS contribute to conversations about HCML. Was HCML a term that could only be applied to CS research, but not more than that?

In short, pinning down a definition for HCML is tricky because it must contend with these questions. In response, I found that many proposed definitions pull HCML to narrow conceptualizations of its scope and impact. Some researchers conceptualize the term around values and operationalize those as guideposts for HCML. The most popular of these values is "fairness" and the closely related concept of "justice". Other scholars focus on methodological plurality, often drawing from qualitative methods (such as interviews and user observation) in HCI, the social sciences, and the humanities. Yet another contingency conceptualizes HCML as applied research areas as human-centered domains, like ML for health. What "counts" as HCML, then, becomes bounded in one of these boxes: values, methods, or domains. But to me, these definitions always fell flat. In focusing on procedures or values, these definitions missed a key part of the whole concept – what it meant to be "human-centered".

I returned to prior work to examine how others have thought about the concept. Some may assume that "human-centered" is a novel term to this technical moment, but in fact it has close precedents to "human-oriented" research in CS and other disciplines. The earliest mention of the concept comes from over 50 years ago in the 1970s when engineers coding computational systems used "human-oriented design" to distinguish themselves from work that just had a technical focus. In so doing, they highlighted the interplay of human values in engineering software systems that supported human labor [25].

The phrase "human-centered" or "human-oriented" is also not exclusively the property of CS and engineering fields. The term has also appeared in information sciences and STS. In information sciences, scholars argued that being "human-centered…must take account of varied social units that structure work and information—organizations and teams, communities and their distinctive social processes and practices" [28]. Perhaps the most straightforward definition was offered by Gasson (2003):

*Human-centered systems production should concern itself with the joint questions of `What can be produced?' and `What should be produced?' The first is about what is technically feasible, the second about what is socially desirable* [21] [p.32]

What stuck out to me in this definition was the understanding of the gap *between* methods and values of what *can* be produced and what *should be* produced, which is a definitely moral call to action. I was struck by the similarity of this definition of Gasson's with Mark Ackerman's "social-technical gap", as proposed in the early days of Computer Supported Cooperative Work (CSCW) [1]. At the time, CSCW was mostly focused on building technology to facilitate collaboration in workplaces. Ackerman pointed out that CSCW must contend with the gap between our social imaginations and the technical realities of what we could engineer. To do so, he argues, we must acknowledge that our systems are "rigid and brittle" about the nuanced social work in which people work. Ackerman suggests ways to manage this gap in the same ways prior work on human-centeredness has approached this problem – considering the ways to "bridge" the social-technical gap by reintroducing theory that recontextualizes CSCW. Bridges are eminently practical in that they provide a path to connect two distant things – in this case, methods and values.

The field of ML now has a similar challenge designing for and bridging social imaginations and values with ML methods that Ackerman describes for CSCW several decades ago. I think of this as a bridge - "human-centered" served as a bridge between values and methods and domains.

Using this prior work as groundwork for discussing practices with HCML, I propose a more unifying definition statement for how we should think about HCML:

*Human-centered machine learning is a set of practices for building, evaluating, deploying, and critiquing ML systems that balances technical innovation with an equivalent focus on human and social concerns.*

HCML is a set of practices that guide better decision-making throughout the whole ML pipeline of problem brainstorming, development, and deployment. It attempts to bridge the difficult world of values with methodological and technical innovation. Our technical innovations are used by others, and the acknowledgment of this can assist in mitigating, avoiding, and even improving outcomes of ML systems. When HCML serves as a dual focus rather than a laundry list of methods or values, we can return to the initial intentions of human-centered thinking as it was introduced in the 70s to CS – a way to approach problem-solving. Therefore, questions from my colleagues that interrogate the edges of the concept of HCML are answered with, "We can look at the details of how something was conceptualized, executed, and evaluated to know if it was HCML."

Why focus on practices? Practices are actionable compared to values because they offer ways to move forward and bridge values into real decisions. I'm not the first to propose that practices are important [33,42]. As mentioned before, Ackerman raised similar questions in CSCW of how to overcome the social-technical gap [1], and Philip Agre did this in the mid-1990s with "critical technical practice" [3]. This article's emphasis on practices also answers Mittelstadt's call for practices that can map abstract ideas like values onto everyday practices [33]. Practices are also embedded in the thought process with these decisions from their start because they directly affect what you do. In these ways, they provide opportunities for reflection for decision-making.

## 3 PRACTICES OF HUMAN-CENTERED MACHINE LEARNING

In this next section, I outline practices that are key to conducting HCML. These practices draw from areas essential to bridging the social and technical realities of HCML research as well as the computational traditions from HCI and AI. The challenges of representing "messy" social contexts in computer code have been a central concern of both of these areas [24], and scholars have spoken to complementary challenges of translating social worlds to the needed abstraction of systems [36,43]. Additionally, I look to ideas from critical perspectives within STS, and critical data studies HCI [3,18,45] and synthesize practices from emerging work that is actively grappling with these problems [22,26,38]. Many of these ideas have been inspired by my experiences working in this space without the HCML label to accompany it.

### 3.1 ENSURE ML IS THE RIGHT SOLUTION AND APPROACH TO TAKE

Due to hype and industry interest, many assume that ML is an appropriate (and often the best) solution for a problem. Some engineers and researchers use ML as a silver bullet, a straightforward solution to complex problems that have resisted other approaches. However, there is an assumption embedded in this popular narrative of ML's success that suggests: because we can use ML, we should. Treating ML as a silver bullet is what Evgeny Mozorov coined as the term "technological solutionism" nearly a decade ago [34], and what Selbst et al. further clarifies as the "solutionism trap," in which CS narratives imply that technology can solve a problem [43]. The first practice, then, is to reframe these traps into an actionable component of HCML problem formulation and conceptualization. For each application, we should ask ourselves "Is ML the right approach to take in the first place?"

Avoiding technological solutionism is more nuanced than just weighing the pros and cons of a potential ML solution. This practice asks us to evaluate whether ML is a good solution and if we have correctly articulated the problem that ML says it solves in the first place. In my work, we have considered building ML systems that could prevent online content moderators from seeing graphic content in mental health postings. This statement presumes that moderators should not see graphic content and that ML could make a meaningful dent in how much graphic content a moderator sees. This solution does not consider that there are other problems in their work—like unsafe working conditions and psychological trauma—that may not be solved by AI alone but in tandem with other changes, such as corporate structure and compensation. In many solutions, the people involved are not asked what an ML solution could look like for them.

What is one way to make progress towards this practice? Jess Holbrook recalls saying to engineers, "Don't expect Machine learning to figure out what problems to solve"[2]. He calls on us to use needs-finding exercises to triangulate what peoples' needs are before engineering energy is spent on developing outcomes for them. ML engineers are not the experts of a "target population", and by assuming we know better, we risk causing more harm than good. At a minimum, we should go ask our target populations what they want and need.

Furthermore, attenuating ourselves to this practice is more than just asking "what are our target population's needs?" because that question doesn't consider if we should use ML at all to address a problem. Nushi et al. describes this as "accountability" of our technical prowess to social outcomes [35]. Considering this question early and often in the thought process may raise the uncomfortable truth that the ML solution should never happen in the first place [6]. We have responsibilities to our systems and the sociotechnical impacts of their design and deployment—the process is not simply data-design-deploy-done.

At its most radical, HCML should consider the possibility that the best way to do this is by refusing ML development. Refusal gives power back to the communities that ML may take agency away from. This refusal might look like manifest-nos[4] of communities that rightfully resist technological intervention as thought of by computer scientists. Similarly, a truly human-centered approach allows communities the space to resist ML and instead advocate for the most just and feminist futures. By including this work inside the space of HCML, we allow for these important conversations to happen rather than relegate them to the sidelines of CS.

### 3.2 ACKNOWLEDGE THAT ML PROBLEM STATEMENTS TAKE POSITIONS

At the point where ML has been chosen as a solution, the next step is formalizing the problem statement that HCML is going to solve. Problem statements are necessary to operationalize abstract ideas into concrete items. A classic framing of ML problem statement is, "given a series of *p* posts in X domain, what is the outcome of the binary prediction task *t*?" Problem statements help teams focus on outcomes, divide up work, and establish parameters for a successful project.

---

[2] https://medium.com/machine-learning-for-all/what-is-human-centred-machine-learning-a2f8f8170f73
[4] https://www.manifestno.com/

However, after deciding that something is a problem and breaking it down into discrete tasks, ML engineers make subtle decisions to accomplish those goals. These decisions will impact how that problem will be instantiated in the world and therefore how data is sampled and labeled, what methods are chosen, and how to evaluate the success of the solution.

I believe that a problem statement for an HCML solution is a *position* statement that reflects peoples' perspectives and assumptions. The idea of positions comes from feminist scholars who recognized that peoples' identity characteristics and social facets, like their gender and race, influenced how people understood and perceived the world [15,37]. To be clear, positionality is more than just acknowledging that men or women have different perspectives on the world – these facets have influenced what and how we make becomes technical reality [16]. For example, in my research, sometimes I frame diagnosis of mental illness as a binary state of being, rather than representing relapse, recovery, or transitions between states of better or worse well-being. In this example, I make assumptions to make the problem statement easier to solve. These decisions have impacts on the downstream system that is built and people whose lives could be influenced by this system.

To illustrate this, I offer an example. ML systems that predict criminality from facial features of mugshots take a *position* on who may be engaging in criminal activity (people with police photographs) and then decide that facial features contain information about criminality. None of this is grounded in immutable truth about the world. Criminality is determined by the legal system under which someone operates, which differs across governments. There is enormous variance in who is arrested for potentially committing crimes, who may have a police photograph taken, and cultural as well as legal considerations of who is eventually found guilty of a crime. This is notwithstanding the highly controversial question of whether cultural traits of criminals can be physiologically detected in facial features. And, in fact, my own social and identity characteristics, like my ethnicity, race, and gender, have implications on how I perceive the criminal justice system and what facial features I see as relevant for prediction if any.[19] These decisions are impacted by the social world we live in, namely the cultural and legal system, not by objective facts about how criminality is defined.

How might this perspective on positions be integrated into an HCML agenda? Positions say, "I see that my experiences of the world impact how I view this problem, and I'm going to document, address, and interrogate them as best I can." This can be in published position statements in papers and documentation about the perspectives the team brings to the task and how these positions informed decision-making. For example, a large body of my research focuses on high-risk mental illnesses, from eating disorders to opioid use disorder and suicidality. I have personal experiences with mental illness, though not with the ones that appear in my work. Driven by my experiences, my position has influenced how I report data and protect participant privacy, consider ethics board review and care of the research, and how I describe the outcomes of studies. The point of positions is not virtue signaling to get published, nor self-flagellation or emotional theater. We all have unique perspectives on how we have experienced the world – it is ok to acknowledge that it impacts how we do our work in HCML. This practice asks us to ask ourselves - who are we as researchers, and what perspectives do we bring to problems that influence our choices?

## 3.3 MOVE BEYOND USERS AND INTERACTION AS THE DEFINITION OF HUMAN

To start any HCML project, it is essential to clarify who matters in our research pursuits. Historically, CS has mapped "humans" to the notion of "use"—a single person's interaction with or relationship to a technology [5]. This concept is useful because it constrains attention on who and what to focus on during design.

ML challenges who a user is and what constitutes interaction. Take the example of "users" of facial recognition technology in airport security. Traditionally, the user is mapped to the person whose face is recognized or the customs agent analyzing the results of this process. However, other individuals that engage with the system aren't captured in traditional interaction metaphors. In the facial recognition example, a person may be in the background and ambiently recognized without directly engaging with or "using" the technology. There are other stakeholders as well, including all the people whose photographs served as training data, with or without their consent, as well as those that refuse to use facial recognition technology.

To avoid these pitfalls and reveal the true diversity of interactions between machine learning and people, HCML must move beyond just an "interaction mode" of understanding users [5]. We must consider multiple ways of engaging with technology—directed and interactive, as well as ambient and non-use—as important questions to engage with for understanding ML systems and broader implications for HCML. Clarifying who is involved can help manage who should be involved in the development lifecycle of a system.

Several existing analytical lenses might facilitate this process for HCML. One lens that has been proposed is the "post-user" lens, which Baumer and Brubaker (2017) define as meaningfully engaging with the complex interactions that people can have with technology. Another approach is an ecosystemic model, or systems-style thinking, in which the considerations of users evolve to understanding stakeholders [23], each with needs and preferences. Ecosystems recognize distributed impacts of people impacted by technology by using their data, ambient use of systems, or decision to not use technological systems at all. Finally, there is inspiration to be taken from the area of needs assessment, asking who are our stakeholders and what are their needs?

## 3.4 LEGITIMIZE DOMAIN CONTRIBUTIONS/COLLABORATIONS AS FIELD PRIORITIES

In addition to prioritizing the needs of stakeholders, we should also look to collaborative fields that have been in conversation with HCML. The true potential of a human-centered perspective considers human and social factors in technological development, and therefore we should consider other fields as equitable contributors to HCML.

Domains outside of machine learning possess the theory, methods, and insights to help HCML build more thoughtful systems. Current research in HCI-AI collaborations points to starting opportunities for value-sensitive approaches in HCML. Examples of new techniques to the area include "human-subject experiments, surveys, and ethnographic inquiries" [47]. Many of these traditions draw from HCI, meaning that this is an easy avenue for initial collaborations and research [4,48], and much work in HCML currently reflects this exchange between HCI and ML. Researchers should be rewarded for engaging in this work, such as producing fairness audits and interpretability tools.

However, interdisciplinarity for HCML raises challenging questions within CS about what constitutes computing research. CS has a longstanding history of using external domains as fertile grounds for advancing its insights. This history includes CS theory being advanced from economics and mechanism design, and neural and genetic algorithms being adapted from biological systems. Classic ML textbooks acknowledge that ML is indebted to an interdisciplinary history of fields like psychology, statistics, and organizational studies – other areas notwithstanding [32]. Like the interdisciplinary histories of AI and HCI—in which building tools that try to resolve technical problems are only one step towards embracing a human-centered agenda—we must start investigating these perspectives in our work by rewarding them.

We need different epistemological approaches to ML that push the applications of ML to other domain areas. There should be more than just a collaboration inside CS with its sub-fields who, given common disciplinary backgrounds, may inadvertently slide into old thought patterns [2]. We should also look to external domains for pieces to the puzzle of our work and invite contributions into our discipline and positions in our departments and research teams [40]. These domains may include critical data studies, sociology, gender and race studies, and organizational studies.

### 3.5 DESIGN HCML ANTICIPATING AND ITERATING ON SOCIOTECHNICAL FAILURE

ML historically prioritizes technical conceptions of performance through quantitative metrics, such as evaluating error rates and efficiency. These metrics are essential in understanding the success or failure of a given system [35].

In addition to the evaluation of metrics for failures, it is becoming obvious that models produce other kinds of failures. These failures are not readily diagnosed with mathematical techniques like evaluating performance. Instead, these failures are embodied within the social and political world that the technical model inhabits. These sociotechnical failures articulate what has been made "marginal" in our designs, where our abstractions and representations of ML do not match reality [2] [p 45]. Consider the challenges of training language systems on African-American Vernacular English [7]. This failure is a breakdown of a system to technically manage different speech patterns, but also a sociotechnical failure of not prioritizing a dialect that is biased against an important ethnic community.

Rather than viewing failure as defeat, what if we were to view failure as a new lens of understanding for how should HCML work? The study of technical and sociotechnical failure is a crucial opportunity for HCML to inspect its sociotechnical impacts on the world, avoiding Selbst et al's notion of "traps" in abstraction [43]. In the study of design, scholars have discussed the notion of sociotechnical failure through metaphors like failures, gaps, brittleness, and breakdowns [1,27]. As a scholar of why technology breaks down (a closely related idea to failure), Steve Jackson argues that "worlds of maintenance and repair and the instances of breakdown… are not separate or alternative to innovation, but sites for some of its most interesting and consequential operations" [27].

An HCML-led study of breakdown and failure should encompass more than how ML operates at a mathematical level, and what it does in the world; it should study the contextual, embedded, and "messy" use of the technology [18]. A solid study of failure, therefore, would attend to not only the WHAT of failure and its cause, but WHO is impacted by failure, in what ways are they impacted by, how they cope with, and operate around failure. Current exemplars of this work are studying the places of AI systems failure [35], and navigation points around assumptions of failure as built into high-stakes medical decision-making [9].

To truly move the needle on HCML, we must go beyond expecting engineers and researchers to ask the intended users and stakeholders what failure may be. Technologists should commit to following their designs into the messiness of the world to see how people grapple with failures, whether they be unintended quirks or more insidious long-term inequalities.

Finally, an anticipatory focus on failure highlights places where technical innovation should never happen in the first place [6], returning to the question posed in Section 3.1. Machine learning cannot solve wicked problems in the world, so rather than shy away from this responsibility, we ought to embrace the messiness in our designs.

## 4 MOVING FORWARD WITH PRACTICES OF HCML

What should we do to see these practices adopted in HCML? Let us look to CS as a field to begin.

For example, to embed the practice of moving beyond HCI notions of the user, research could start by using stakeholder analysis and needs assessment to guide value-sensitive algorithm design [48]. When deciding if an ML solution is appropriate, interdisciplinarity calls us to read external literature and identify collaborations. From there, a decision may be reached about whether the costs of the ML system are worth it. More humanistic positions can be identified; for example, if classifying mental illness has problematic outcomes on how we define or diagnose illness, ML may not be worth applying in these scenarios or we may need to adjust our training data.

However, these practices also influence our thinking about disciplines writ large. One recent example is NeurIPS, which requires that authors must disclose the negative implications of design. This effort follows the ACM Future of Computing Academy's recent blog post arguing for similar practices[7]. These considerations of potential implications directly advance the study of sociotechnical failure being normalized in the community. This move by NeurIPS echoes the call of Winograd and Flores (1986) to open "the space of possibilities for actions" of the engineer to engage with their creations [46] [165]. To see the space of possibility, I encourage readers to reflect on the practices mentioned above in their work throughout the pipeline of ML.

Similarly, academic institutions can support these practices through administrative efforts like cross-appointments and creating interdisciplinary funding opportunities. At higher levels, HCML might be driven through reorganization around concepts like "computing" rather than just "computer science," or incentivized through formal mechanisms like paper publication standards and review guidance like the example of NeurIPS provided earlier. Pragmatic work across teams that incorporates human-centered design and evaluation perspectives [31] could also open collaborative opportunities in industry. Grant opportunities that center considerations of these principles alongside traditional calls would be another way to further entrench them in this area.

Practices are everyday yet instrumental to how work is conducted. Therefore, I consider HCML to be driven by practices more than just a checklist of methods or self-stated values. By applying these practices, we are better able to direct our decisions, which in turn inform meta-level considerations for the field of computing moving forward.

---

[7] https://acm-fca.org/2018/03/29/negativeimpacts/